\documentclass{bmvc2k}
\usepackage{graphicx}
\usepackage{amsmath,amssymb} 
\usepackage{amsmath,lipsum}
\usepackage{wrapfig}
\usepackage[export]{adjustbox}
\usepackage{booktabs}
\usepackage{algorithm}
\usepackage{algorithmic}
 \usepackage{array,multirow,graphicx}
 \usepackage{float}
\usepackage{bbold}

\title{G$^2$NetPL: Generic Game-Theoretic Network for Partial-Label Image Classification}

\addauthor{Rabab Abdelfattah}{rabab@email.sc.edu}{1}
\addauthor{Xin Zhang}{xz8@email.sc.edu}{1}
\addauthor{Mostafa M. Fouda}{mfouda@ieee.org}{2}
\addauthor{Xiaofeng Wang}{wangxi@cec.sc.edu}{1}
\addauthor{Song Wang}{songwang@cec.sc.edu}{3}

\addinstitution{
Department of Electrical Engineering\\
University of South Carolina\\
South Carolina, USA
}
\addinstitution{
Department of Electrical and Computer Engineering\\
Idaho State University\\
Idaho, USA
}
\addinstitution{
Department of Computer Science\\
University of South Carolina\\
South Carolina, USA
}
\runninghead{ABDELFATTAH, ET AL.}{G$^2$NetPL for Partial-Label Image Classification}

\def\etal{\emph{et al}\bmvaOneDot}

\begin{document}

\maketitle
\vspace{-0.35 cm}
\begin{abstract}
Multi-label image classification aims to predict all possible labels in an image. It is usually formulated as a partial-label learning problem, since it could be expensive in practice to annotate all the labels in every training image. Existing works on partial-label learning focus on the case where each training image is labeled with only a subset of its positive/negative labels.  To effectively address partial-label classification, this paper proposes an end-to-end Generic Game-theoretic Network (${\rm G^2NetPL}$) for partial-label learning, which can be applied to most partial-label settings, including a very challenging, but annotation-efficient case where only a subset of the training images are labeled, each with only one positive label, while the rest of the training images remain unlabeled.
In ${\rm G^2NetPL}$, each unobserved label is associated with a soft pseudo label, which, together with the network, formulates a two-player non-zero-sum non-cooperative game.  The objective of the network is to minimize the loss function with given pseudo labels, while the pseudo labels will seek convergence to 1 (positive) or 0 (negative) with a penalty of deviating from the predicted labels determined by the network.  In addition, we introduce a confidence-aware scheduler into the loss of the network to adaptively perform easy-to-hard learning for different labels.  Extensive experiments demonstrate that our proposed ${\rm G^2NetPL}$ outperforms many state-of-the-art multi-label classification methods under various partial-label settings on three different datasets.

\end{abstract}

\vspace{-0.35 cm}
\section{Introduction}
\vspace{-0.35 cm}
Based on deep-learning techniques, significant progress has been made on single-label image classification~\cite{deng2009imagenet} where each image only has one label. However, in many real applications, one image may contain multiple objects and/or exhibit multiple attributes which cannot be well described by a single label. This leads to an important computer-vision task of multi-label image classification that aims to identify all the labels of an image. One main challenge in deep-learning based multi-label image classification comes from the requirement of large-scale labeled training images. In particular, many supervised-learning algorithms require all the labels present in each training image to be accurately annotated and this full-label manual annotation can be very difficult and laborious~\cite{deng2014scalable}. 

To relieve the annotation burden of full labeling, several recent works on multi-label classification consider the use of training image data with partial labels~\cite{cole2021multi,durand2019learning,huynh2020interactive,kundu2020exploiting,xu2013speedup,PinedaSalvador2019im2set,ishida2017learning}, where only a limited number of labels are annotated on each image.  A special case is the full-set-single-positive-label (FSPL) setting, which requires to annotate only one positive label for each training image~\cite{cole2021multi}. While these partial-label settings can mitigate the annotation burden, one still needs to label all training images. To go one step further, the subset-single-positive-label (SSPL) setting only requires to label a subset of the training images and for each labeled image, only one positive label is annotated \cite{abdelfattah2022plmcl}. \textcolor{black}{For example, each image in the iNaturalist dataset only has one positive label with other classes unlabeled~\cite{cole2021multi}. Adding new unlabeled images to enrich this dataset leads to SSPL, which is considered a practical scenario. Actually, this applies to most datasets for multi-class classification. Adding new unlabeled images makes those datasets suitable for the multi-label problem under SSPL without annotation costs.}

Clearly, this SSPL setting can further reduce the annotation burden for large-scale datasets. 
However, the inclusion of unlabeled images raises a great challenge in network design and training.  On one hand, those existing partial-label learning methods based on label correlation~\cite{huynh2020interactive} and label matrix completion~\cite{cabral2011matrix} are not applicable to SSPL since they cannot handle single positive labels and unlabeled data simultaneously. On the other hand, although semi-supervised learning can deal with unlabeled images, it is usually designed based on the assumption of having a subset of fully-labeled training images, but not specifically for SSPL setting that does not offer such a subset.  Therefore, the performance of the existing semi-supervised models on SSPL setting may be degraded. 

This paper presents a new generic game-theoretic network (${\rm G^2NetPL}$) for end-to-end training of the multi-label classifier under SSPL setting. In ${\rm G^2NetPL}$, each unobserved label is associated with a soft pseudo label, which acts as a player in a two-player non-zero-sum non-cooperative game with the objective of converging to 1 (positive) or 0 (negative), given a penalty of deviating from the predicted labels determined by the network. As the second player, the network is to minimize a weighted loss function with given pseudo labels. 
Our contributions are summarized as follows:
\begin{itemize}
\vspace{-0.25cm}
\item 
We propose a novel ${\rm G^2NetPL}$ for partial-label image classification, focusing on SSPL setting with the understanding that ${\rm G^2NetPL}$ can be applied to most, if not all, partial-label settings. Within this game-theoretic framework, the two players, the network and the pseudo labels, are updated with different objectives.  Accordingly, different loss functions are developed for players. Therefore, the network can, from a game-theoretic point of view, achieve robustness with respect to errors in pseudo label estimation.  Meantime, the existence of Nash equilibrium will guarantee convergence in training.  
\vspace{-0.25cm}
\item
We introduce the confidence level of pseudo labels into our loss function, where less weights will be considered for those pseudo labels with low confidence levels.  This is critical, especially at the early stage of training when the observed labels are lack and the confidence levels on the pseudo labels are low. Otherwise, premature convergence could be achieved with low confidence levels due to lack of observed labels.
\vspace{-0.25cm}
\item
Extensive experimental results show that ${\rm G^2NetPL}$ outperforms the state-of-the-arts under different partial-label settings on three widely-used datasets. With fewer observed labels in SSPL setting, our method can still get comparable classification results as those methods using FSPL setting.
\end{itemize}

\vspace{-0.55cm}
\section{Related Work}
\label{sec:related}
 \vspace{-0.23cm}
This section briefly reviews previous works for partial multi-label image classification.

\smallskip
\noindent{\bf Partial Labels.} 
To address the unobserved labels, which can be positives or negatives, several works simply tend to regard unobserved labels as negative ~\cite{sun2017revisiting,mahajan2018exploring,bucak2011multi,chen2013fast}. This assumption usually leads to a significant performance drop by incorrectly initializing the positive labels as negatives~\cite{joulin2016learning}. Others tend to find the unobserved labels via label correlations~\cite{mahajan2018exploring,wu2015ml}, label matrix completion~\cite{wang2014binary,cabral2011matrix}, low-rank learning \cite{yu2014large,yang2016improving}, and probabilistic models \cite{kapoor2012multilabel,chu2018deep}. Most of these methods do not allow end-to-end training.
Recently, several end-to-end models are proposed for the partial-label setting~\cite{cole2021multi,durand2019learning,huynh2020interactive}. In~\cite{huynh2020interactive}, image and label similarity graphs are built to estimate the unobserved labels. The construction of label graphs relies on the label co-occurrence information in the training data.  This method is not applicable to single-positive-label (SPL) settings which has at most one observed label per image. 
Durand \etal~\cite{durand2019learning} adopt both the graph neural network and curriculum learning to find the relations between the labels and complete the unobserved labels. This approach depends on a fixed threshold to add “easy-unobserved” labels to the loss function.  Once the network is updated, those previously predicted unobserved labels will be discarded, which makes this approach heuristic without continuity in learning. Different from that,  ${\rm G^2NetPL}$ can forecast convergence and optimality.  In addition, our model will use every unobserved labels in training, while in~\cite{durand2019learning} there is a possibility that some unobserved labels are never selected before the model converges, due to the threshold-based~strategy.

Semi-supervised learning, on the other hand, is a widely used technique for leveraging a large unlabeled dataset alongside a small fully-labeled subset~\cite{wang2018adaptive,niu2019multi,liu2006semi,sohn2020fixmatch,berthelot2019mixmatch,rizve2021in}. As mentioned in \cite{rizve2021in}, however, most semi-supervised models focus only on single-label classification problem.
To fit into multi-label classification, modifications are necessary.
Other related works exploit the label ranking \cite{kanehira2016multi} and label correlations  \cite{sun2010multi} to learn from positive and unlabeled data. However, these methods require more than one positive label per image so that pairwise label dependencies can be utilized. 

\smallskip
\noindent{\bf Full-set Single Positive Label.} 
The FSPL setting was proposed by Cole \etal~\cite{cole2021multi} most recently, where only a single positive label is observed on each training image.  This model jointly trains the image classifier and the label estimator using the online label estimation. However, the online estimation requires to store the full labels of all the training data in the memory \cite{cole2021multi}. In addition, the label estimator is randomly initialized which leads to inferior model in end-to-end setting and the observed/unobserved labels are addressed with equal weights.  Different from that, ${\rm G^2NetPL}$ defines a game between the network and the pseudo labels, each of which has its own goal to achieve.  Moreover, the introduction of the confidence level places different weights for unobserved labels in the loss function of the network, which can gradually move the learning from observed labels to unobserved labels without being over-confident. Moreover, ${\rm G^2NetPL}$ learns pseudo labels for images in each mini-batch without requiring keeping all training data in the memory.

\smallskip
\noindent{\bf Subset Single Positive Label.} 
To the best of our knowledge, there is no work explicitly exploring SSPL in-depth for multi-label image classification. As mentioned above, with necessary modifications, some semi-supervised learning algorithms, such as Fixmatch~\cite{sohn2020fixmatch} and UPS~\cite{rizve2021in}, can work under this setting.  In the experiments, we compare the performance of our proposed ${\rm G^2NetPL}$ with these semi-supervised methods
under SSPL settings.

\vspace{-0.35 cm}
\section{The ${\rm G^2NetPL}$ Framework}
\vspace{-0.35cm}
\textbf{Notations.}
Let $\mathcal I$ be the set of all training images and $\mathcal D$ be the subset of $\mathcal I$, of which each element has a single positive label. The images in $\mathcal I\setminus \mathcal D$ remain unlabeled. Let $y_o^i$ be the $L$-dimension label vector associated with image $i \in \mathcal D$, where $L$ is the number of the total classes. The $j$th entry of $y_o^i$ can be 1 or $\emptyset$, which means that the $j$th class is observed positive or unlabeled, respectively. Therefore, $\mathcal D$ provides the ground truth characterized in $y_o^i$. We denote the classifier by $f(i,\theta_t)$ that maps image $i$ into a predicted label vector $\hat{y}^i_t\in [0,1]^L$, i.e., $\hat y^i_t = f(i, \theta_t)$. Here $\theta_t$ is the network parameters obtained at epoch~$t$ during training. To characterize the unobserved labels during training, we take advantage of soft pseudo labels~\cite{tanaka2018joint} and use $\hat{y}^i_{u,t} \in [0,1]^L$ to denote the soft pseudo label vector of  image $i \in \mathcal I$ at epoch $t$.  
In the following discussion, we will drop the index $i$ for notational simplicity, if it is clear in context.
The cross-entropy loss function between two scalars $p,q \in [0,1]$ is defined as
$\mathcal{L}(p,q)=-p\log(q)-(1-p)\log(1-q)$. 
Given $v\in \mathbb R^L$, $[v]_j$ denotes the $j$th entry of $v$.
\begin{figure*}[!t]
  \centering
  \includegraphics[scale=0.38]{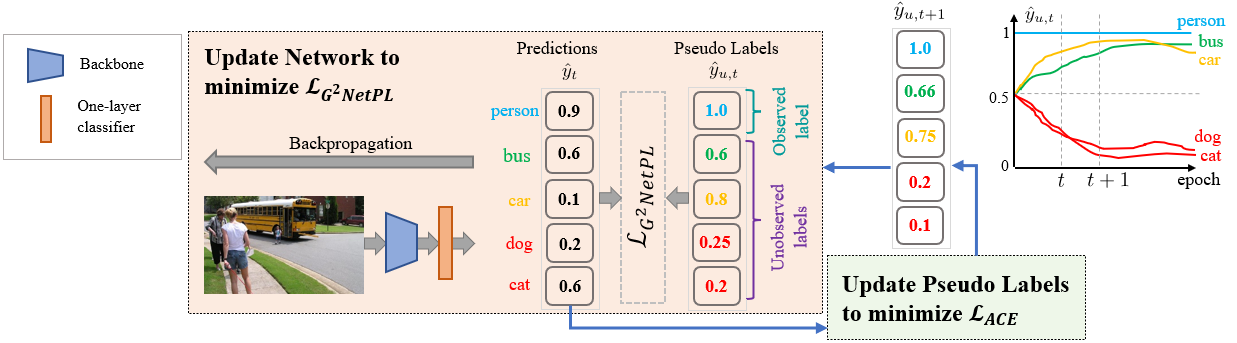}
  \vspace{-0.25 cm}
  \caption{An overview of ${\rm G^2NetPL}$ that consists of two players: the network and the pseudo labels.  The network includes backbone and one-layer classifier to generate a prediction for an input image. The loss function for the network is $\mathcal L_{\rm G^2NetPL}$ that is a function of the network parameters $\theta_t$ with given pseudo labels.  The pseudo labels $\hat y_{u,t}$ are iterated to minimize the cross-entropy loss function $\mathcal L_{\rm ACE}$ between the pseudo labels and the predicted labels. }
  \label{fig:G2Net}
  \vspace{-0.4 cm}
\end{figure*}
\vspace{-0.25 cm}

\vspace{-0.15 cm}
\subsection{Overview}
\vspace{-0.25 cm}
The ${\rm G^2NetPL}$ is shown in Fig.~\ref{fig:G2Net}, which introduces a two-player non-zero-sum non-cooperative game between the network and the pseudo labels. \textcolor{black}{The 2-player game we employ contains three basic elements of a game: a set of players (pseudo labels and network), a set of strategies or actions for players ($y_u$ for pseudo labels and $\theta$ for the network), and the associated costs ($\mathcal L_{\rm ACE}$ and $\mathcal L_{\rm G^2NetPL}$). In our game, the payoff of each player is determined by the strategies of both players.  The objective of the network is to minimize the loss function $\mathcal L_{\rm G^2NetPL}$ ($y_u$ affects $\mathcal L_{\rm G^2NetPL}$ through the scheduler $\xi$), while the goal of the pseudo labels is to converge to 1 or 0 with a penalty of deviating from the predicted labels generated by the network ($\theta$ affects $\mathcal L_{\rm CE}$ through the predicted labels).  The Nash equilibrium makes both players have no incentive to change its current strategy.} 
During training, the pseudo label vector $\hat{y}_{u,0}$ of each image in~$\mathcal I$ will be initialized with 1 for the observed labels and the unbiased probability $0.5$ for the unobserved labels, which means equal distance to 1 (positive) and 0~(negative). Based on received $\hat y_{u,t}$, the network updates its parameter $\theta_t$ which minimizes the loss function $\mathcal L_{\rm G^2NetPL}$.  Once the $\theta_t$ is determined, the pseudo label $\hat y_{u,t+1}$ will be iterated to minimize cross-entropy loss function $\mathcal L_{\rm ACE}$, given $\hat y_{t}$ predicted by the network with $\theta_t$.  This process repeats until both network parameters and the pseudo labels converge to Nash equilibrium, which provides robustness in learning.  During testing, only the backbone and the classifier will be used to predict the labels.  
\vspace{-0.25 cm}
\subsection{The Loss of Pseudo Labels}
\vspace{-0.25 cm}
As mentioned in context, the objective of the pseudo labels is to converge to the true 1 or 0 with the predicted labels from the network as a reference.  A penalty will be placed when the pseudo label $\hat y_{u,t}$ deviates from the predicted label $\hat y_{t}$.  To present the loss function $\mathcal L_{\rm ACE}$, we first introduce the pseudo label latent parameter $y_{u,t}\in \mathbb R^L$ and a mapping function $F:\mathbb R \to \mathbb [0,1]$ such that
\vspace{-0.25 cm}
\begin{equation}
    [\hat y_{u,t}]_j = F([y_{u,t}]_j), ~ j=1,2,\cdots,L.
\end{equation}
The pseudo label latent parameter $y_{u,t}$ is actually a project of $\hat y_{u,t}$ on $\mathbb R^L$.  
Inversely, the function $F(\cdot)$ regulates $[y_{u,t}]_j$ into $[0,1]$.  We define the augmented cross-entropy loss $\mathcal L_{\rm ACE}$ based on $y_{u,t}$, adding a penalty on $[\hat y_{u,t}]_j$ when deviating from 0 and 1.  For instance, the loss function to be minimized can be formulated as 
\vspace{-0.1 cm}
\begin{align}
\mathcal L_{\rm ACE}(\hat y_t,{y}_{u,t}) = \sum_{j=1}^L \left[\mathcal L\left([\hat y_{t}]_j,F([{y}_{u,t}]_j]\right) + \lambda_j F([y_{u,t}]_j) (1-F([y_{u,t}]_j)) \right]
\end{align}
where $\lambda_j$ is a positive weighting constant.  Notice that the function $F(\cdot)$ must be a function which outputs 1 when $[y_{u,t}]_j$ approaches $+\infty$ and $0$ when $[y_{u,t}]_j$ approaches $-\infty$.  If $F$ is smooth, then based on the chain rule in derivative,
\begin{equation}
\nabla \mathcal {L_{\rm ACE}}_{[y_{u,t}]_j}(\hat y_t,{y}_{u,t})  = \left(\frac{[\hat y_{u,t}]_j - [\hat y_t]_j}{[\hat y_{u,t}]_j (1-[\hat y_{u,t}]_j) } + \lambda_j - 2 \lambda_j [\hat y_{u,t}]_j \right) \cdot F'([y_{u,t}]_j).
 \end{equation}

\begin{wrapfigure}{h}{4.7cm}
  \centering
  \vspace{-5.5mm}
  	\includegraphics[scale=0.20]{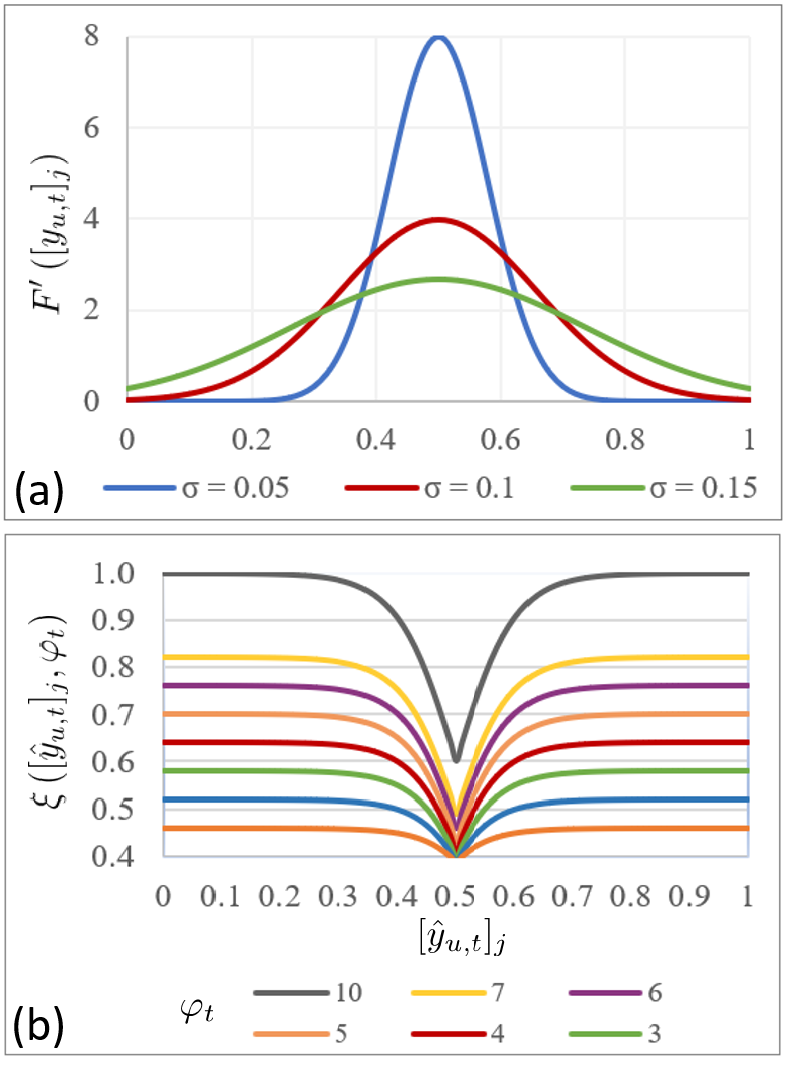}\\
\vspace{-0.25 cm}
  \caption{(a) The probability density function of different Gaussian distributions; (b) The scheduler $\xi([\hat y_{u,t}]_j,\varphi_t)$.}
  \label{fig:parameters}
  \vspace{-0.35 cm}
\end{wrapfigure}
Letting $\nabla \mathcal {L_{\rm ACE}}_{[y_{u,t}]_j}(\hat y_t,{y}_{u,t})=0$ indicates that the optimal $[y_{u,t}]_j$ will satisfy either $F'([y_{u,t}]_j) = 0$ or $\frac{[\hat y_{u,t}]_j - [\hat y_t]_j}{[\hat y_{u,t}]_j (1-[\hat y_{u,t}]_j) }+ \lambda_j - 2 \lambda_j [\hat y_{u,t}]_j =0$.  If we can obtain an explicit solution on $[\hat y_{u,t}]_j)$ in the former equation, then the computation cost can be dramatically saved.  Meantime, the choice of $\lambda_j$ can be time-varying depending on $[\hat y_t]_j$, i.e. $\lambda_j =\lambda([\hat y_t]_j)$.  If $[\hat y_t]_j$ is close to 0.5, which means unreliable, $\lambda_j$ can be large; otherwise, it can be small, such as the curves indicating in Fig.~\ref{fig:parameters}(a).
The choice of $F(\cdot)$ is not unique and will affect the performance of the classifier. For instance, one candidate is the sigmoid function.  Another interesting option is the cumulative distribution function (CDF) of a Gaussian distribution with the mean at 0.5, where the associated probability density function is
\vspace{-0.25 cm}
\begin{align}
    F' ([{y}_{u,t}]_j) = \frac{1}{\sigma \sqrt{2\pi}} e^{- \frac{1}{2} \left(\frac{[{y}_{u,t}]_j-0.5}{\sigma}\right)^2},
    \label{eq:psi}
\end{align}
as shown in Fig.~\ref{fig:parameters}(a).
In this case, $F' ([{y}_{u,t}]_j) = 0$ indicates that $[{y}_{u,t}]_j = \pm \infty$, which are the generalized roots to $\nabla \mathcal {L_{\rm ACE}}_{y_{u,t}}$, corresponding to $[\hat {y}_{u,t}]_j = 1$ (positive) or $0$ (negative). The parameter $\sigma$ is the standard deviation of this Gaussian distribution.  It determines how fast the probability increases from 0 to 1.  For instance, when $[{y}_{u,t}]_j=0.5$, $F ([{y}_{u,t}]_j)$ achieves its maximal increasing rate, $F' (0.5) = \frac{1}{\sigma \sqrt{2\pi}}$.  The smaller $\sigma$ is, the faster $F([{y}_{u,t}]_j)$ changes from 0 to 1.  In an extreme case when $\sigma=0$, $F' (0.5) = + \infty$, then $F$ will be degraded to a step function with 0.5 delay, which implies hard pseudo labels, i.e. $[\hat {y}_{u,t}]_j =1$ if $[{y}_{u,t}]_j \ge 0.5$ and $[\hat {y}_{u,t}]_j =0$ if $[{y}_{u,t}]_j < 0.5$.  Since using hard pseudo labels may lose smoothness of $\mathcal {L_{\rm ACE}}$ and make the results over-confident~\cite{szegedy2016rethinking}, it suggests that $\sigma$ cannot be extremely small.  On the other hand, $\sigma$ cannot be too large.  Otherwise, $F'([{y}_{u,t}]_j)$ will be small.  It means that $F([{y}_{u,t}]_j)$ is not anxious to push $[{y}_{u,t}]_j$ to 1 or 0 at all, which could be contracted with the objective of the pseudo labels that seek convergence to 1 or 0. Finally, the choice of the augmented loss function is not unique either.  For instance, another candidate can be 
\vspace{-0.18 cm}
\begin{align}
\mathcal L_{\rm ACE}(\hat y_t,{y}_{u,t}) = \sum_{j=1}^L  e^{\lambda_j F([y_{u,t}]_j) (1-F([y_{u,t}]_j))}\mathcal L\left([\hat y_{t}]_j,F([{y}_{u,t}]_j]\right),
     \vspace{-2 cm}
\end{align}
which will amplify the cross-entropy loss when $[y_{u,t}]_j$ is far away from 0 or 1.  When the loss function becomes complicated, numerical methods are necessary for the solutions.  

\vspace{-0.25 cm}
\subsection{Scheduled Loss of the Network}
\label{sub:loss-function}
To learn the parameters of the primary multi-label CNN, the following loss function must be minimized
\vspace{-0.2 cm}
\begin{equation}
    \mathcal{L}_{\rm G^2NetPL} = \mathcal{L}_{\rm obs} + \mathcal{L}_{\rm unobs}.
    \label{eq:loss}
\end{equation}
$\mathcal{L}_{\rm obs}$ is the binary cross-entropy classification loss between the observed labels and the associated predicted labels plus the regularizer that implicitly penalizes false positives (more details on the regularizer can be found in~\cite{cole2021multi}).  $\mathcal{L}_{\rm unobs}$ is the weighted cross-entropy classification loss between the pseudo labels for the unobserved labels and their associated predicted labels.  Let $\mathcal U_i$ be the set of the unobserved labels in image~$i$.  Then $\mathcal{L}_{\rm unobs}$ is defined as
\begin{align}
    \mathcal{L}_{\rm unobs} = \sum_{i \in \mathcal I} \sum_{j\in \mathcal U_i} \xi([\hat y_{u,t}]_j,\varphi_t) \mathcal L\left([\hat y_{t}]_j,[\hat{y}_{u,t}]_j\right)
     \vspace{-5 cm}
\end{align} 
where $\xi([\hat y_{u,t}]_j,\varphi_t)$ is the confidence-aware scheduler that controls the metric learning loss and $\varphi_t$ is the current epoch $t$ divided by the total number of epochs (representing a linear scheduler~\cite{wang2019dynamic}).
In $\mathcal{L}_{\rm unobs}$, we exploit the soft pseudo labels instead of sharp $1$ or $0$ since the latter may push the predictions
to be over-confident~\cite{szegedy2016rethinking}. In contrast, using the pseudo labels as the distribution scores may inherit the idea of label-smoothing regularization~\cite{szegedy2016rethinking}. 
The purpose of introducing the scheduler is to weight the unobserved loss.  Different timed schedulers have been developed such as dropout training~\cite{morerio2017curriculum}, transfer learning~\cite{weinshall2018curriculum}, and self-correction~\cite{li2020self}, in which cases the weighted score only relies on the iteration steps (or epochs). Practically, if the scheduler depends on both the confidence and the iteration steps, it may lead to improved performance.  For instance, when the pseudo label $[\hat y_{u,t}]_j$ has low confidence, the related loss $\mathcal L\left([\hat y_{t}]_j,[\hat{y}_{u,t}]_j\right)$ may deviate far away from the actual loss between the predicted label and the true label (although unknown). Therefore, the scheduler $\xi([\hat y_{u,t}]_j,\varphi_t)$ must be small to reduce the impact of such deviation on the total loss $\mathcal L_{\rm G^2NetPL}$. On the contrary, if $[\hat y_{u,t}]_j$ has high confidence, $\xi([\hat y_{u,t}]_j,\varphi_t)$ should be large. The pseudo labels will gradually build up their confidence during iterations.  Following this idea, we introduce a new confidence-aware scheduler as follows, inspired by~\cite{wang2020progressive}:
\vspace{-0.22 cm}
\begin{align}
    &\xi([\hat y_{u,t}]_j,\varphi_t)  =\beta \frac{1-\gamma e^{-10 |2[\hat{y}_{u,t}]_j-1|} }{1+ \gamma e^{-10|2[\hat{y}_{u,t}]_j-1|} } + (1-\beta) \varphi_t,
    \label{eq:sched}
     \vspace{-1 cm}
\end{align}
where $\beta, \gamma$ are the positive hyper-parameters.
Notice that the confidence of a pseudo label is reflected by the term $|2\hat{y}_{u,t}-1|$. For instance, if $[\hat{y}_{u,t}]_j$ is 0.5, the related $\xi$ will achieve its minimum, which means that the network is not confident about the pseudo label and the related loss should contribute less in $\mathcal L_{\rm G^2NetPL}$.  Otherwise, if $[\hat{y}_{u,t}]_j = 0$ or $[\hat{y}_{u,t}]_j = 1$, the related $\xi$ reaches its maximum, which indicates high confidence on the current pseudo label and the associated loss should contribute more. 
In Fig.~\ref{fig:parameters}, the scheduler has lower weighted scores over the interval $0.3 \le [\hat{y}_{u,t}]_j \le 0.7$, because it represents the region with lower confidence and the results generated over this region may potentially degrade the performance.

\vspace{-3mm}
\begin{algorithm}[ht]\small
	\caption{${\rm G^2NetPL}$ Training}
		\begin{algorithmic}[1]
		\REQUIRE  input image $i \in \mathcal{I}$;
		\REQUIRE observed label $y_{o}$;
		\REQUIRE  neural network $f(i, \theta)$ with parameters $\theta$ and input $i$.
		\REQUIRE 
        $[\hat{y}_{u}]_j = 1 \mbox{ or } 0$,~ if labeled; 
        $[\hat{y}_{u}]_j = 0.5$,~ if unlabeled\\
		\STATE \textbf{Repeat}
		\begin{ALC@g}
		\STATE $\xi \longleftarrow \beta \frac{1-\gamma e^{-10 |2[\hat{y}_{u}]_j-1|} }{1+ \gamma e^{-10|2[\hat{y}_{u}]_j-1|} } + (1-\beta) \varphi$;
		 \STATE $\theta \longleftarrow    \arg\min_{\theta}\mathcal L_{\rm G^2NetPL}$ (through back-propagation);
    		\STATE $\hat{y} \longleftarrow f(i, \theta)$;
    	    \STATE $y_{u} \longleftarrow \arg \min_{y_u} \mathcal L_{\rm ACE}(\hat y,{y}_{u})$; 
    	    \STATE $[\hat{y}_{u}]_j \longleftarrow F([y_{u}]_j)$;
	    \end{ALC@g}
        \STATE \textbf{until} the max iteration or convergence;
        \STATE \textbf{Output:} $\theta$, pseudo labels $\hat{y}_u$, and predicted labels $\hat y$.
	\end{algorithmic}
\end{algorithm} 
\vspace{-0.25cm}

\vspace{0.1 cm}
\noindent
\textbf{Existence and Convergence of Nash equilibrium.}
According to the Debreu-Glicksberg-Fan results~\cite{debreu1952social,fan1952fixed}, if every player's strategy set is compact and convex, and the payoff functions are all continuous and concave (or convex if minimizing the cost), then a pure strategy Nash equilibrium exists. In our case, the pseudo label $\hat y_u$ stays inside $[0,1]$ which is compact and convex.  The related loss function $\mathcal L_{\rm ACE}$ is convex w.r.t. $\hat y_u$. Meantime, ResNet has been shown to be (near-)convex in~\cite{park2022vision}.  Therefore, as long as the network parameters stay in a compact and convex set (although this set could be very large), Nash equilibrium will exist and convergence can be achieved using our proposed algorithm  based on fixed-point theorem~\cite{kakutani1941generalization}. 

\vspace{0.1 cm}
\noindent
\textcolor{black}{\textbf{Advantages of our pseudo labels compared to existing methods.}
\textit{(i)} Our approach can ensure smoothness and convergence of pseudo labels during training, which are important quality indices, while most existing methods (e.g., FixMatch~\cite{sohn2020fixmatch} and UPS~\cite{rizve2021in}) use threshold-based
strategies that lack the continuity and guarantee of convergence in pseudo label updating. \textit{(ii)} Our pseudo labels, as a player in the game, have their own goal, independent of the network loss, which provide a second opinion on the quality of the network training to avoid over-fitting, while in most existing work, pseudo labels either do not have a clear goal or share the same loss as the network. \textit{(iii)} All pseudo labels in our model will contribute to the network loss except weight differences, while in threshold-based strategies, there is a chance that some pseudo labels are never selected for network training, which may result in information loss.}

\vspace{-0.45 cm}
\section{Experiments}
\vspace{-0.25 cm}
\noindent
\textbf{Datasets.} The \textbf{PASCAL VOC} \cite{everingham2011pascal} consists of 20 classes in 5,717 training images and the results are reported on the official validation set (5,823 images). \textbf{MS-COCO}~\cite{lin2014microsoft} consists of 80 classes, including 82,081 training images and 40,137 testing images. The \textbf{NUS-WIDE} \cite{chua2009nus} contains 81 classes and it is split into 150K for training and 60.2K for testing.\\
\noindent
\textbf{Implementation Details.} 
We use the same backbone architecture and the same classifier, ResNet-50~\cite{he2016deep} pre-trained on ImageNet~\cite{russakovsky2015imagenet}, for all comparisons. The training is performed in the end-to-end setting, where the parameters of the backbone and the classifier are updated and
trained for 10 epochs~\cite{cole2021multi}. The learning rate is used, varying from $10^{-2}$ to $10^{-5}$, with the batch size in {$8$, $16$}. The best mean average precision (mAP) on the validation set is reported. 

\vspace{-0.3 cm}
\subsection{Full-set Single Positive Label (FSPL)}
\label{subsub:FSPL}
\vspace{-0.15 cm}
\begin{wraptable}{!t}{8.0cm}
\vspace{-0.46 cm}
\caption{Quantitative results (mAP) of multi-label image classification on four different datasets. 
Bold represents the highest mAP and underline represents the second-best among FSPL setting (Single positive and No negative).}
\vspace{0.05 cm}
\centering{
\begin{adjustbox}{width=0.6\textwidth}  
\begin{tabular}{|l|c|c||c|c|c|}
\hline
&\multicolumn{2}{c||}{\textbf{Observed}}&\multicolumn{3}{c|}{\textbf{End-to-End Setting}}\\
\cline{2-6}
Losses&Positive&Negative&VOC&COCO&NUS\\
\hline
$\mathcal{L}_{\rm BCE}$\cite{nam2014large}&All&All& 89.1&75.5&52.6\\
$\mathcal{L}_{\rm BCE-LS}$&All& All& 90.0&76.8&53.5\\
\hline
$\mathcal{L}_{\rm IU}$ \cite{durand2019learning}&Single & Single& 83.2&59.7&42.9\\
\hline
$\mathcal{L}_{\rm AN}$ \cite{kundu2020exploiting}&Single& No & 85.1&64.1&42.0\\
$\mathcal{L}_{\rm AN-LS}$ \cite{cole2021multi}&Single& No &86.7&\underline{66.9}&44.9\\
$\mathcal{L}_{\rm WAN}$ \cite{mac2019presence}&Single& No  & 86.5&64.8&\underline{46.3}\\
$\mathcal{L}_{\rm EPR}$~\cite{cole2021multi}&Single& No  & 85.5&63.3&46.0\\
$\mathcal{L}_{\rm ROLE}$~\cite{cole2021multi}&Single& No  & \underline{87.9}&66.3&43.1\\
\textbf{$\mathcal{L}_{\rm G^2NetPL}$ (ours)}&Single& No& \textbf{88.8}&\textbf{72.4} & \textbf{49.7}\\
\hline

\end{tabular}
\end{adjustbox}
}
\label{tab:FSL}
\vspace{-0.42 cm}
\end{wraptable} 
We first conduct experiments with FSPL setting which has $100\%$ of the training images annotated with only one single positive label.  To do so, we follow~\cite{cole2021multi} by assigning randomly only one positive label for each image in the training set with the guarantee that each class has at least one image labeled.  

\smallskip
\noindent
\textbf{Baseline.}   
Several existing multi-label approaches are discussed in~\cite{cole2021multi}, based on the variants of the binary cross-entropy (BCE) loss which is widely used in multi-label classification.
The compared approaches in our experiments include:   ignoring all unobserved labels $\mathcal{L}_{\rm IU}$, assuming the unobserved labels are negatives $\mathcal{L}_{\rm AN}$~\cite{kundu2020exploiting}, smoothing the former loss function~$\mathcal{L}_{\rm AN-LS}$, down-weighting the terms in the loss function related to negative labels~$\mathcal{L}_{\rm WAN}$~\cite{mac2019presence}, expected positive regularization $\mathcal{L}_{\rm EPR}$~\cite{cole2021multi}, and online estimation of unobserved labels~$\mathcal{L}_{\rm ROLE}$~\cite{cole2021multi}. 
The evaluation under the FSPL setting is divided into three categories in Table~\ref{tab:FSL}. In the first category, the results are evaluated in the case where all positive and all negative labels are available (``All'') with the binary cross-entropy loss function $\mathcal{L}_{\rm BCE}$ and smoothing $\mathcal{L}_{\rm BCE-LS}$. In the second category, the results are evaluated in the case of assigning single observed label, positive and negative ("Single"). The results in these two categories provide a reference of the learning performance when all labels, or more labels, can be observed.  The third category employs a fair comparison of approaches under FSPL setting.  ${\rm G^2NetPL}$ runs under this category.

\smallskip
\noindent
\textbf{Discussion.} 
As reported in Table~\ref{tab:FSL}, ${\rm G^2NetPL}$ is trained under FSPL setting, but still achieves comparable results to $\mathcal{L}_{\rm BCE}$ and $\mathcal{L}_{\rm BCE-LS}$, which rely on fully observed labels on VOC, COCO, and NUS datasets.
It means that ${\rm G^2NetPL}$ requires much fewer labels without significantly sacrificing the precision. For instance, on the COCO dataset, ${\rm G^2NetPL}$ only uses 1.25\% of the total labels to train the model, while the performance is only dropped by 4.4\%, compared with the baseline (second row in Table~\ref{tab:FSL}) that are trained with 100\% of the labels. In the third category of FSPL setting, ${\rm G^2NetPL}$ exceeds the highest scores from the existing methods by 0.9\%, 5.5\% and 3.4\% over VOC, COCO, and NUS datasets, respectively. 

\begin{figure}[!t]
  \centering
  	\includegraphics[scale=0.35]{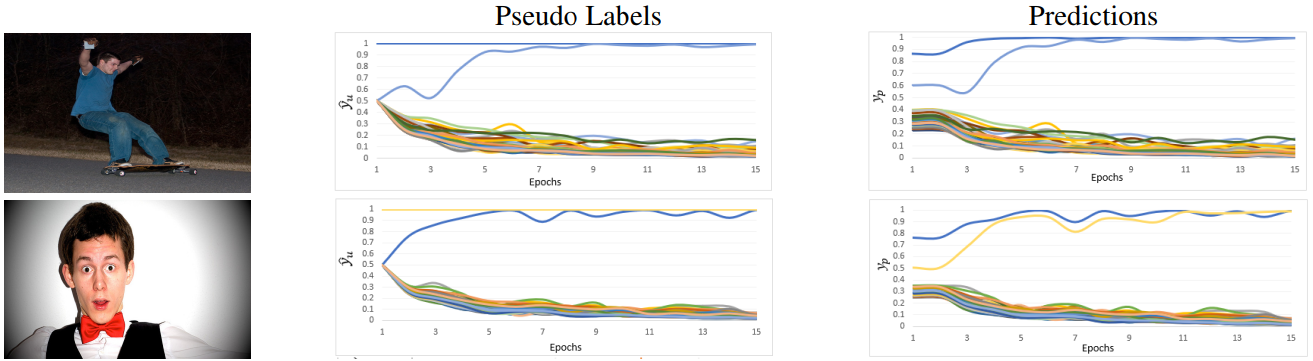}\\
  	  \vspace{-0.25 cm}
  \caption{Convergence for pseudo labels and the predictions in COCO dataset for FSPL.}
  \label{fig:conv}
  \vspace{-0.3 cm}
\end{figure}

\noindent
\textcolor{black}{\textbf{Quality of Pseudo Labels.}The quality can be evaluated by two metrics. (i) The closer a pseudo label is to 0 or 1, the more
indicative it is, which means higher quality. This principle is carried by the scheduler in ${\rm G^2NetPL}$ during training.
(ii) If a pseudo label oscillates a lot over epochs, it may be
unreliable. So the quality can be evaluated by smoothness
and convergence, while this can be ensured by our model. Fig. \ref{fig:conv} shows that our pseudo labels converge without much oscillation. We evaluate the
final pseudo labels against the ground truth (GT) on Pascal under FSPL. The mAP is 93.6\% for ours and 89.8\% for
ROLE \cite{cole2021multi}, which shows our pseudo labels are closer to GT.}
\vspace{-0.4 cm}
\subsection{Subset Single Positive Label (SSPL)}
\label{subsub:SSPL}
\vspace{-0.19 cm}
\begin{table*}[!t]\small
\caption{Quantitative results (mAP) of multi-label image classification on different subsets of the single observed label (SSPL) setting on three different datasets with End-to-End learning settings. Bold represents the highest mAP and underline represents the second-best.}
\label{tab:SSPL}
\vspace{-0.02 cm}
\centering{
\begin{adjustbox}{width=0.95\textwidth}
\begin{tabular}{|l|l|l|l|l||l|l|l|l||l|l|l|l|l|l|l|}

\hline
Losses&\multicolumn{4}{c||}{COCO}&\multicolumn{4}{c||}{VOC}&\multicolumn{4}{c|}{NUS}\\
\cline{2-13}
&20\%&40\%&60\%&80\%&20\%&40\%&60\%&80\%&20\%&40\%&60\%&80\%\\
\hline
$\mathcal{L}_{\rm AN}$ \cite{kundu2020exploiting} &46.3&53.8&59.5&62.4&51.3&71.7&80.2&82.8&28.5&35.2&38.9&40.8\\
$\mathcal{L}_{\rm AN-LS}$  &48.9&57.9&62.3&\underline{65.5}&70.0&79.0&85.0&86.1&27.8&36.0&39.1&41.4\\
$\mathcal{L}_{\rm WAN}$ \cite{mac2019presence} & \underline{57.0}&\underline{60.9}&\underline{63.5}&64.5&\underline{76.4}&\underline{82.5}&85.1&85.6&\underline{37.6}&\underline{41.3}&\underline{43.7}&\underline{44.8}\\
$\mathcal{L}_{\rm EPR}$ \cite{cole2021multi} &52.7&58.4&61.5&62.6&75.5&81.0&83.9&84.5&34.9&39.5&42.3&44.2\\
$\mathcal{L}_{\rm ROLE}$ \cite{cole2021multi} &47.3&57.6&62.7&65.2&66.9&80.9&\underline{85.9}&\underline{86.8}&27.0&32.2&37.1&39.7\\

\textbf{$\mathcal{L}_{\rm G^2NetPL}$ (ours)} &\textbf{62.2}&\textbf{65.8}&\textbf{69.7}&\textbf{71.2}&\textbf{79.6}&\textbf{85.2}&\textbf{87.6}&\textbf{88.2}&\textbf{38.4}&\textbf{42.9}&\textbf{46.9}&\textbf{48.5}\\
\hline 
\end{tabular} \end{adjustbox}}
\vspace{-0.3 cm}
\end{table*}
The comparison results of SSPL setting are reported in Table~\ref{tab:SSPL}, where $20\%$ in the table means that only $20\%$ of the total images have a single positive label and the rest of images are totally unlabeled. 

\smallskip
\noindent
\textbf{Discussion.} According to Table~\ref{tab:SSPL}, ${\rm G^2NetPL}$ outperforms all the compared methods with different percentage of observed labels over all datasets. 
Among these compared approaches, $\mathcal{L}_{\rm AN}$ is considered the simplest which is widely explored \cite{joulin2016learning,kundu2020exploiting,mahajan2018exploring}. However, its performance is degraded by assuming the unobserved labels as negatives since those false negatives will become noisy labels. Using label smoothing with $\mathcal{L}_{\rm AN}$ gives $\mathcal{L}_{\rm AN-LS}$, which can overcome the negative impact of noisy labels in the multi-class setting and achieve reasonable performance as compared to~$\mathcal{L}_{\rm AN}$. As suggested in~\cite{cole2021multi}, we use $\mathcal{L}_{\rm AN-LS}$ as a baseline for our work. 
We also observe that increasing the number of unobserved labels did not significantly affect the performance of $\mathcal{L}_{\rm WAN}$ because down-weighing the terms related to the negative labels may help reduce the effect of noisy labels.  
$\mathcal{L}_{\rm EPR}$follows the same conclusion and performs results very close to $\mathcal{L}_{\rm WAN}$, since the former ignores the unobserved terms and focuses on finding the positive labels according to its regularizer. 
Finally, we notice that $\mathcal{L}_{\rm ROLE}$ shows a significant drop when the subset of the labeled images is small (i.e. 20\%) in the end-to-end setting, due to random initialization of the estimated labels~\cite{cole2021multi} as well as equally weighting the observed and unobserved labels. 
The ${\rm G^2NetPL}$ still outperforms the other methods since our game-theoretic framework not only avoids the negative factors mentioned before, but augments the label confidence into the weights as well.    
In fact, in 60\% and 80\% SSPL settings, the scores of our method (reported in Table~\ref{tab:SSPL}) are even comparable to that under the FSPL setting (reported in Table~\ref{tab:FSL}). According to \textbf{SSPL versus FSPL,} we observed that ${\rm G^2NetPL}$ under 60\% and 80\% SSPL settings (Table~\ref{tab:SSPL}) achieves comparable results to ROLE under FSPL setting (Table \ref{tab:FSL}). 



\vspace{-0.4 cm}
\subsection{Compared to Semi-Supervised Models.} 
\label{subsec:SSM}
\begin{figure}[!t]
  \centering
  	\includegraphics[scale=0.39]{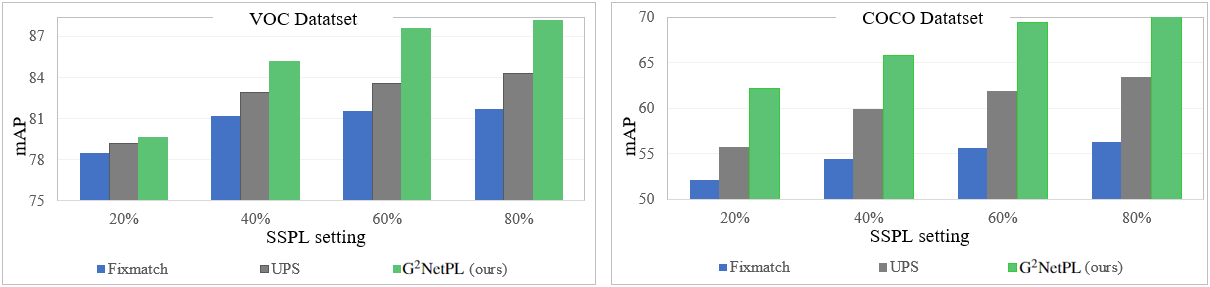}\\
  	  \vspace{-0.25 cm}
  \caption{Comparison between our proposed model and semi-supervised models on SSPL.}
  \label{fig:semi-super}
  \vspace{-0.4 cm}
\end{figure}

\vspace{-0.25 cm}
This subsection compares the performance of ${\rm G^2NetPL}$ with the state-of-the-art semi-supervised models that exploit the pseudo labels to train the network, such as FixMatch~\cite{sohn2020fixmatch} and UPS~\cite{rizve2021in}, on two datasets VoC and COCO under different SSPL settings. 
\noindent
As shown in Fig.~\ref{fig:semi-super}, ${\rm G^2NetPL}$ outperforms FixMatch and UPS in both datasets.  This is because both FixMatch and UPS models suffer from the same issue of using threshold-based strategies as~\cite{durand2019learning} (more details can be found in Section~\ref{sec:related}.  There is a lack of continuity in pseudo label learning. In addition, a considerable number of unobserved data may be ignored during the entire training. On the contrary, ${\rm G^2NetPL}$ (game-theoretic framework) enables continuous pseudo label training and therefore is able to capture the temporal information more effectively.

\vspace{-0.3 cm}
\subsection{Ablation Study} 
\label{sub:ablation}
\vspace{-0.18 cm}
\noindent
\textbf{${\bf G^2NetPL}$ Modules.}   
The results are reported in Table~\ref{tab:ablation1} for SSPL (20\%) and FSPL with linear-init setting. Linear-init learning trains the classifier linearly for 25 epochs with fixed parameters of the backbone and then fine-tunes for another 5 epochs to update the parameters of both the classifier and the backbone~\cite{cole2021multi}.   
We start with the basic variation of the model which uses the cross-entropy loss on the observed labels. In this case, we assume that all the unobserved labels are negative, since ignoring the unobserved labels may lead to overfitting.
When adding components into the model, this assumption can be removed since we use the pseudo labels to deal with the unobserved labels.  We first use the sigmoid function as $F(\cdot)$ to update the soft pseudo labels which improve the mAP under different settings and datasets. Adding the confidence-aware scheduler and applying the cumulative distribution function of Guassian distribution as $F(\cdot)$ further enhances the overall performance, since they work together to make the training adaptive with respect to the confidence of the pseudo labels. 
\begin{table}[!h]\small
\vspace{-0.15 cm}
\caption{Ablation study for all proposed model components on SSPL (20\%) and FSPL.}
\vspace{-0.05 cm}
\label{tab:ablation1}
\centering{
\begin{adjustbox}{width=0.95\textwidth} 
\begin{tabular}
        {
      |
      >{}p{0.3\textwidth}|
      >{\centering}p{0.16\textwidth}|
       >{\centering}p{0.16\textwidth}|
        >{\centering}p{0.16\textwidth}|
      >{\centering\arraybackslash}p{0.16\textwidth}
      |
    }
\hline
&\multicolumn{2}{c|}{COCO dataset}&\multicolumn{2}{c|}{NUS dataset}\\
\cline{2-5}
${\rm G^2NetPL}$ Modules&  SSPL (mAP)&FSPL (mAP)&  SSPL (mAP)&FSPL (mAP)\\
\hline
Baseline&56.6&67.3&36.0&46.8\\
Use sigmoid as $F(\cdot)$&62.4&71.1&44.5&49.9\\
Add the scheduler $\xi$&64.4&71.4&44.8&50.4\\
Use CDF in \eqref{eq:psi} as $F(\cdot)$&65.0&72.7&45.3&50.9\\
\hline
\end{tabular}
\end{adjustbox}
}
\vspace{-0.4 cm}
\end{table}

\vspace{-0.35 cm}
\section{Conclusions}
\vspace{-0.4 cm}
This paper presented a novel ${\rm G^2NetPL}$ for multi-label image classification under partial-label settings. In ${\rm G^2NetPL}$, two players, the network and the pseudo labels, formulate a non-zero sum game, where the network minimizes the network loss $\mathcal L_{\rm G^2NetPL}$ and the pseudo labels minimize $\mathcal L_{\rm ACE}$ that reflects the idea of converging to 1 or 0 but penalizing deviation from the predicted labels. Through extensive experiments on different partial-label settings, we demonstrated that our ${\rm G^2NetPL}$ outperforms several state-of-the-art methods by training on fewer observed labels. ${\rm G^2NetPL}$ is generic to all kinds of partial-label settings. 

\vspace{0.45 cm}
\noindent
\textbf{Acknowledgements.}
The authors gratefully acknowledge the partial financial support of the National Science Foundation (1830512 and 2018966).

\bibliography{egbib}
\end{document}